\documentclass[a4paper]{scrartcl}

\usepackage[utf8]{inputenc}
\usepackage[T1]{fontenc}
\usepackage[english]{babel}
\usepackage{amsmath,amssymb,amsfonts,siunitx,commath}
\usepackage{amstext}
\usepackage{mathtools, bbm}
\usepackage{amsthm}
\usepackage{tikz,url}
\usepackage{geometry}
\usepackage{mathtools}
\mathtoolsset{showonlyrefs}
\usepackage{subcaption}

\usepackage{soul}
\usepackage{color}

\usepackage[section]{algorithm}
\usepackage{algorithmic}
\usepackage{hyperref}
\usepackage{enumerate}
\usepackage{listings}
\usepackage{footmisc}

\DeclareMathOperator*{\argmin}{arg\,min}

\newcommand{\R}{\mathbb{R}}

\newcommand{\tT}{\mathrm{T}}

\theoremstyle{plain}
\newtheorem{lemma}{Lemma}

\newtheorem{remark}[lemma]{Remark}

\theoremstyle{definition}

\renewcommand*{\phi}{\varphi}
\renewcommand*{\rho}{\varrho}

\renewcommand*{\subset}{\subseteq}

\renewcommand*{\norm}[1]{\ensuremath{\left\Vert #1 \right\Vert}}

\newcommand*{\bfv}{\ensuremath{\mathbf{v}}}

\begin{document}
\title{Lagrangian Motion Magnification with Double Sparse Optical Flow Decomposition}
	\author{
Philipp Flotho$^{2}$
\and
Cosmas Heiss$^{1}$
\and
Gabriele Steidl$^{1}$
\and
Daniel J. Strauss$^{2}$
}
\date{\today}

\maketitle

\footnotetext[1]{
TU Berlin,
Stra{\ss}e des 17. Juni 136,  
D-10587 Berlin, Germany,
\{steidl\}@math.tu-berlin.de.
} 
\footnotetext[2]{
Systems Neuroscience and Neurotechnology Unit, Neurocenter, Faculty of Medicine, Saarland University \& School of Engineering, htw saar and Center for Digital Neurotechnologies Saar (CDNS), Germany,
\{Philipp.Flotho, Daniel.Strauss\}@uni-saarland.de.
} 

\begin{abstract}
Microexpressions are fast and spatially small facial expressions that are difficult to detect. Therefore
motion magnification techniques, which aim at amplifying and hence revealing subtle motion in videos,
appear useful for handling such expressions.
There are basically two main approaches, namely via Eulerian or Lagrangian techniques. 
While the first one magnifies motion implicitly  by operating directly on image pixels,
the Lagrangian approach uses optical flow (OF) techniques to extract and {magnify} pixel trajectories. 
In this paper, we propose a novel approach for local Lagrangian motion magnification of facial micro-motions.
Our contribution is three-fold: 
first, we fine tune the recurrent all-pairs field transforms (RAFT) for OFs 
deep learning approach for faces by adding ground truth obtained from the variational 
dense inverse search (DIS) for OF algorithm applied to the CASME II video set of facial micro expressions.
This enables us to produce OFs of facial videos in an efficient and sufficiently accurate way.
Second, since facial micro-motions are both local in space and time, we propose to approximate the OF field
by sparse components both in space and time leading to a double sparse decomposition.
Third, we use this decomposition to magnify micro-motions in specific areas of the face,
where we introduce a new forward warping strategy using a triangular splitting of the image grid and
barycentric interpolation of the RGB vectors at the corners of the transformed triangles.
We demonstrate {the feasibility} of our approach by various examples.
\end{abstract}

%------------------------------------------------------------
\section{Introduction} \label{sec:intro}
%------------------------------------------------------------
Motion magnification describes a wide variety of algorithms to {magnify} and therefore visualize subtle, imperceptibly small motions in videos. 
In analogy to fluid dynamics, motion magnification techniques can be grouped into Eulerian and Lagrangian approaches. 

Eulerian methods  have first been introduced by Wu et al. \cite{wu2012eulerian}.
These methods amplify motion or color changes by modifying  the time-dependent color variation in the video  using  temporal or spatial filtering, e.g., by dealing with
image pyramids of different resolutions \cite{elgharib2015video,  wadhwa2013phase, wu2012eulerian, zhang2017video}. The term Eulerian emphasizes that information is processed on the fixed grid of pixels while Lagrangian methods manipulate point trajectories.  Eulerian methods found applications for quantitative motion assessment as well.  For example, Sarrafi et al.  \cite{sarrafi2018vibration} and Eitner et al. \cite{eitner2021modal} 
used point tracking on motion magnified recordings for the detection and modal parameter estimation of vibrations in different mechanical components. 
Fei et al. \cite{fei2021exposing} applied motion magnification to the detection of AI-generated videos and
Alinovi et al. \cite{alinovi2018respiratory} to respiratory rate monitoring.
While Eulerian methods are often very easy and fast to evaluate and can therefore be readily employed in real-time applications, the type and degree of motions that can be {magnified} is limited. 
In particular, phase-based methods depend on  local spatial frequencies which can lead to blurring around sharp images edges. 
Intensity-based Eulerian methods magnify motion by linearization,  which is known to be limited to small movements.

Lagrangian approaches rely on the explicitly computed motion field of objects within the video which is in general done by means {of} OF estimation \cite{ bai2012selectively, FirstMotionMag, flotho2018lagrangian}. 
The OF fields enable not only to selectively and adaptively magnify types of motions that vary spatially as, e.g. in \cite{flotho2018lagrangian,FirstMotionMag},
but also to attenuate them separately or completely remove all movement, see \cite{bai2012selectively, flotho2022a}. An advantage of Lagrangian methods is that the estimated motion information can not only be used for the motion magnification task, but allows subsequent quantitative assessment \cite{hartmann2021measurement, strauss2020vestigial}. 

The advancements in OF estimation in the past decades resulted in high-accuracy methods with different model invariants as well as robustness and high computing speed. Variational methods optimize functions with explicitly modeled data terms and regularizers. They still produce state-of-the-art accuracy in many applications, especially in combination with advanced smoothness terms such as \cite{hartmann2021measurement} and methods for the initialization of very large displacements: Chen et al. \cite{chen2013large} lead (as of March 2022) the Middlebury OF benchmark \cite{baker2011database} with respect to average endpoint error and average angular error. They use similarity transformations for a segmented flow field as initialization of large motion. In a second step, the variational method of Sun et al. \cite{sun2010secrets} is used for subpixel refinement. The widely available dense inverse search (DIS) OF method \cite{kroeger2016fast} initializes the OF field with patch correspondences and uses a variational method similar to Sun et al. and Brox et al. \cite{brox2004high} for high resolution refinement. The availability of large scale synthetic datasets such as MPI Sintel dataset \cite{mayer2016large} paved the way for deep learning-based methods such as RAFT \cite{teed2020raft} which computes a low resolution flow field from correlation volumes which is subsequently upsampled via learned upsampling convolutions. Variants of the method are still among high performing methods in the competitive Sintel benchmark\footnote{http://sintel.is.tue.mpg.de/results} \cite{Butler:ECCV:2012}. 
{Its} fast convergence {\cite{sun2022makes}} makes it in particular interesting for refinement tasks. 

Non-contact monitoring of the affective and neurocognitive state is an emerging topic with many new applications in different areas of life sciences and engineering. Especially, estimation of the psycho-physiological state can find applications in healthcare settings and even neuro-ergonomics in human machine interaction. In that context, facial microexpressions promise to allow an assessment of emotions from faces. In contrast to normal facial expressions, microexpressions {can even} occur when {the person tries to conceal the emotion \cite{yan2013fast}}. 

However, microexpressions are temporally and spatially very small movements and, therefore, difficult to observe. Recent work shows that  small movements, which last less than $\SI{500}{\ms}$ can be considered as microexpressions \cite{yan2013fast}.
{Besides appearance-based features that analyse the skin texture of faces,} OF field-based features have long been used for the detection and analysis of microexpressions \cite{huang2017discriminative,xu2017microexpression,le2018micro,zhang2021review}. { Apart from the small, microexpression induced movements,  appearance-based methods use face appearance features that describe skin texture of faces to capture changes in shading and texture for microexpression detection, see \cite{huang2017discriminative}}.

On a more general scope, micromovements of the face and head can contain other important physiological and psychophysiological information: the movement of the head can contain cues on heart rate \cite{shan2013video}, movement of the mouth can be applied for audio-visual speech recognition in noisy environments \cite{mroueh2015deep},
and micromovements of the ear can contain cues on audio stimuli \cite{strauss2020vestigial}. 
While many of the analysis methods fall back on OF methods for analysis, there are not many annotated datasets for specific, local movements. The selective detection and {magnification} of micromovements of the face is a relatively untapped area of research with the potential to guide the annotation of such datasets and visualize the isolated movements that can be exploited for computer vision algorithms. 

However, such underlying emotional implications will be not in the focus of this paper.
Instead, we will concentrate on the relatively untapped area of detecting and selectively {magnifying} micromovements.
This exceeds the field of microexpressions and goes more in the direction of realistic magnifications in computer graphics.

In this paper, we rely on Lagrangian motion magnification, where
our  approach is based on three pillars. 
First, we fine tune the RAFT deep learning method \cite{teed2020raft} 
using variational approaches \cite{brox2004high, kroeger2016fast} on a facial data set to create new ground truth data to fine tune the network on recordings of faces with small displacements.
Next, we employ ideas of sparse matrix factorization from \cite{sparseDictLearning}, to decompose the OF field into meaningful sparse components. 
Variational approaches with sparsity promoting regularization terms are very common in inverse problems in imaging,
see also \cite{BS2021,CTCB2013,CLMW2011,HK2014} for other sparsity based approaches than just sparse matrix factorization. 
Here we address the sparsity  both in space and time leading to a double sparse representation. Although there exists a broad literature on sparse decomposition, also in connection with principle component analysis, see e.g. \cite{BS2021} and the references therein, 
to the best of our knowledge, such decomposition was not applied for motion magnification so far.
Having the above decomposition at hand,
we are finally able to magnify the motion in respective areas in an unsupervised way. 
For this, we use a sophisticated forward warping method.
Image warping methods, usually simpler backward warping approaches, 
are used in the context of OF estimation, compression or image registration \cite{brox2004high,wohlberg}. 
We will explain forward and backward warping to illustrate why the first one is better suited for our purposes.

The outline of this paper is as follows:
in Section \ref{sec:optical_flow}, we recall the RAFT approach for computing OF fields from deep learning 
and describe our fine-tuning based on a facial dataset and a variational inspired approach.
Then, in Section \ref{sec:PCA}, we show how OF fields from micro-motions can be decomposed sparsely in space and
time. 
Having such a decomposition, we describe how to magnify the motion in the respective areas in 
Section \ref{sec:MM}. In particular, we have to use a specially designed method for forward warping.
Experiments of our method are given in Section \ref{sec:numerics}.
Conclusions and directions of further research are addressed in Section \ref{sec:conclusions}. 

Our software and video results are available at \url{https://github.com/phflot/dsd_momag}.

%------------------------------------------------------------
\section{Optical Flow Detection in Facial Micro-Motions} \label{sec:optical_flow}
%------------------------------------------------------------
In this section, we describe our path for OF detection. 
Then, in the next sections, the motion field will be decomposed and partially magnified.

\paragraph{OF computation by a variational model.}
Our Lagrangian motion magnification approach 
uses the rather accurate variational detection of the OF 
between image frames in a video. 
This is indeed highly nontrivial, since facial micro-motions are in general 
highly local both in space and time and are therefore often imperceptible by a human observer.
OF models for videos are usually based on a gray-value constancy assumption,
meaning that in a continuous model each pixel keeps its value when moving over the time, i.e.,
$$
f(t,x_1(t),x_2(t)) = \text{const} 
$$
and thus
$$
\frac{d}{dt} f = 
\frac{\partial f}{\partial t}  + \frac{\partial f}{\partial x_1} \, \dot x_1(t)+ \frac{\partial f}{\partial x_2} \, \dot x_2(t) = 0.
$$  
Linearization $x_i(t) \approx x_i(t_0) + (t-t_0)v_i(t_0)$ with $v_i(t_0) \coloneqq \dot x_i(t_0)$, $i=1,2$
leads at time $t_0$ to the following underdetermined system for computing the OF vectors $(v_1,v_2)$:
\begin{equation}\label{cba}
\frac{\partial f}{\partial t}  + \frac{\partial f}{\partial x_1} \, v_1(t_0)+ \frac{\partial f}{\partial x_2} \, v_2 (t_0) \approx 0.
\end{equation}
Clearly, digital video sequences are discrete both in space and time.
Given 
$$f(t,x_1,x_2), \quad x_i = 1,\ldots,n_i, \, i=1,2, \; t=1,\ldots,T+1,$$ 
we are interested in the OF fields between \emph{consecutive} video frames
\begin{equation} \label{OF}
\mathbf{v}(t,x_1,x_2) 
\coloneqq 
\begin{pmatrix} 
v_1(t,x_1,x_2)\\
v_2(t,x_1,x_2)
\end{pmatrix}, \quad x_i =1,\ldots,n_i, \, i=1,2,\; t=1,\ldots,T.
\end{equation}
Then the derivatives in \eqref{cba} are replaced
by finite differences and the approximation \eqref{cba} is used 
to form the data term $D(f,v)$ 
within a variational model 
$$
\argmin_v \{D(f,v) + \lambda R(v)\} , \quad \lambda >0.
$$
Here $R(v)$ is a regularizing term or prior of the OF field which is necessary to make the underdetermined problem 
well-posed. 
Different data and regularization terms were used in the literature.
Starting with the Horn-Schunck model \cite{HS81},
there exist meanwhile many sophisticated variational OF models for certain purposes.  
For an overview, we refer to \cite{BePeSch15}. 

\paragraph{OF detection by RAFT.}
While recently introduced deep-learning based OF techniques can outperform variational OF methods in terms of accuracy on many modern benchmarks, their performance depends not only on architectural innovations but also on the quality of the training data and training strategies.
In this paper, we rely on the RAFT approach from \cite{teed2020raft} which demonstrated state-of-the-art performance. 
Training and inference of the original RAFT is briefly described in the next remark.

\begin{remark}(Original RAFT)
First, two consecutive image frames 
$f_1, f_2 \in \R^{n_1,n_2}$  are subsampled to one eighth of their resolution
resulting in $\tilde f_1, \tilde f_2 \in \R^{\tilde n_1,\tilde n_2}$, where $\tilde n_i = n_i/8$, $i=1,2$.
Then RAFT is a deep-learning based model
which computes for two consecutive image frames $\tilde f_1, \tilde f_2$ and for an initial OF field $\tilde v$
an update $\Delta \tilde v$ of the field
\begin{align*}
    \Phi \colon 		\R^{\tilde n_1, \tilde n_2} \times \R^{\tilde n_1, \tilde n_2} \times \R^{2,\tilde n_1, \tilde n_2} 
		\to \R^{2, \tilde n_1, \tilde n_2},\quad
    (\tilde f_1, \tilde f_2, \tilde v) &\mapsto \Delta \tilde v.
\end{align*}
For an initial flow equal to zero, the method realizes, for $r=0,\ldots,R-1$, the following iteration:
\begin{align*}
    \tilde v_0 &:= 0,\\
    \Delta \tilde v_r &:= \Phi(\tilde f_1, \tilde f_2, \tilde v_r),\\
    \tilde v_{r+1} &:=  \tilde v_r + \Delta \tilde v_r.
\end{align*}
For inference, the authors use the output flow field after a fixed number $R$ of such iterations. 
For the $r$-th iterates, learned upsampling convolutions are applied
to bring the computed flow back to the original image resolution, i.e.  
$\tilde v_r \mapsto v_r$. 
Then, for training this model, the authors use the loss function
\begin{align*}
    \mathcal{E} = \sum_{r=1}^R \alpha^{R-r} \|v^{(\text{gt})} - v_r\|_1,
\end{align*}
where $v^{(\text{gt})}$ denotes the ground truth and $\alpha=0.8$ is chosen.
\end{remark}

RAFT was improved with optimized training strategies, datasets and augmentations in various works. 
For a comparison of different methods see \cite{sun2022makes}.
More precisely, the RAFT training schedule  
combines now different datasets in multiple training stages for pre-training and refinement as follows \cite{teed2020raft}: 
RAFT is pre-trained on FlyingChairs \cite{DFIB15} for 100k iterations, 
which is followed by 100k iterations on FlyingThings3D \cite{baker2011database}. 
Further, RAFT is fine tuned on 100k iterations on Sintel \cite{mayer2016large} (RAFT-Sintel), KITTI-2015 \cite{menze2015object} and HD1k \cite{kondermann2016hci}, which is finally followed by 50k fine tuning on KITTI-2015 (RAFT-KITTI).

\paragraph{Fine tuning of RAFT with data produced by variational OF.}
We  fine tune RAFT with additional training data obtained via variational OF methods. To this end, 
we replace the fine tuning on KITTI-2015 with a fine tuning 
on facial microexpression videos derived from the CASME II \cite{yan2014casme}
and SMIC datasets \cite{li2013spontaneous}.
First, we construct a microexpression dataset (ME) by annotating SMIC and CASME II 
with the  dense inverse search optimization (DISO) approach from \cite{kroeger2016fast} implemented 
in the OpenCV library \cite{opencv_library}.
 Note that DISO uses a variational OF technique close to 
\cite{brox2004high,sun2010secrets}.
Second, we enrich the ME dataset as follows:
we use the already estimated OF $\mathbf{v} (t, \cdot)$
and apply it to the first frame $f(1, \cdot)$ of each sequence in SMIC and CASME II
to obtain new sequences  $\hat{f}(t, \cdot + \mathbf{v}(t,\cdot)) := f(1, \cdot)$.
The reason for including both sequences $f$ and $\hat{f}$ in ME is that
\begin{itemize}
\item[i)] for $\hat{f}$, an accurate ground truth displacement field exists, 
but it does not contain the changes in face appearance from the original video, because the color information is just propagated from the first frame of each sequence;
\item[ii)] on the other hand, $f$ contains changes in face appearance from the original video, but there is only a DISO estimation of the OF ground truth.  
\end{itemize}
In summary, with this strategy, we ensure that 
the dataset contains accurate image and flow pairs with $\hat{f}$ 
as well as the original image sequence with dynamic face appearance information in $f$ 
for which we have OF estimates. Including the dynamic face appearance information is particularly important { in the context of microexpressions \cite{huang2017discriminative}}.

%------------------------------------------------------------
\section{Motion Decomposition}\label{sec:PCA}
%------------------------------------------------------------
Once the OF field is known, we want to decompose it into sparse components
in space and time to detect the local facial regions of interest.
Sparsity driven decomposition methods have received a lot of interest in recent years
and there is an overwhelming amount of literature on the topic, 
see \cite{sparseDictLearning} and the references therein.
Our approach is based on an appropriate application and  modification of a method in \cite{sparseDictLearning} 
for our setting.

Facial micro-motions are sparse with respect to space and time.
Let $n \coloneqq n_1 n_2$ be the number of image grid points.
Therefore we aim to decompose the flow field \eqref{OF}
into $K \ll \min\{T,n\}$ components $G^k$ and $\mathbf{d}^k$, 
so that
$$
\mathbf{v}(t,x_1,x_2) \approx \sum_{k=1}^K G^k (x_1,x_2) \mathbf{d}^k (t) , 
\quad 
\mathbf{d}^k (t) \coloneqq 
\begin{pmatrix} 
d_1^k(t)\\
d_2^k(t)
\end{pmatrix}.
$$
The right-hand side separates space and time variables.
On the one hand, this decomposition can be seen for each fixed spatial point evolving 
over the time $t=1,\ldots,T$ as
\begin{align}\label{decomp}
\underbrace{
\begin{pmatrix} 
\mathbf{v}(1,\cdot)\\
\vdots\\
\mathbf{v}(T,\cdot)
\end{pmatrix}
}_{V(\cdot)}
\approx
\underbrace{
\begin{pmatrix} 
\mathbf{d}^1(1) & \ldots &\mathbf{d}^K (1)\\
\vdots          & \vdots & \vdots\\
\mathbf{d}^1(T) & \ldots &\mathbf{d}^K (T)
\end{pmatrix}
}_{D = (d^1| \ldots|d^K)}
\underbrace{
\begin{pmatrix} 
G^1(\cdot)\\
\vdots\\
G^K(\cdot)
\end{pmatrix}}_{G(\cdot)},
\end{align}
and we search for a dictionary or (non-orthogonal) principal components consisting
of the columns of $D$ such that the sample vectors 
$(\mathbf{v}(1,\cdot)^\tT, \ldots, \mathbf{v}(T,\cdot)^\tT) \in \R^{2T}$
at each spatial position can be sparsely represented with respect to this basis.
On the other hand, we can consider at each fixed time all spatial points and the corresponding columnwise reshaped vectors in 
$\R^{n}$, i.e.,
$$
\left( v_i(\cdot,x_1,x_2) \right)_{x_1,x_2}
\approx
\underbrace{\left( \left( G^1(x_1,x_2) \right)_{x_1,x_2} |\ldots | \left( G^K(x_1,x_2)  \right)_{x_1,x_2}\right)}_{G^\tT \, \in \, \R^{n,K}}
\begin{pmatrix}
d_i^1(\cdot)\\
\vdots\\
d_i^K(\cdot)
\end{pmatrix}, \quad i=1,2
$$
and find a dictionary or (non-orthogonal) principal components consisting of the column of $G$ such that the samples of vectors
$\left( v_i(\cdot,x_1,x_2) \right)_{x_1,x_2} \in \R^{n}$, $i=1,2$
at each time can be sparsely represented with respect to this basis.

In summary, we are looking for a doubly sparse model with respect to space and time.
The sparse spatial components $G^k$, $k=1,\ldots,K$ will be later used to {magnify} the motion 
in special regions, while the sparse time components show the duration at which the magnification should appear.
Using the notation in \eqref{decomp}, we propose to find such double sparse decomposition 
by solving for appropriately chosen $\alpha,\beta >0$ the minimization problem
\begin{align} \label{eq:model} 
\argmin_{D,G} \sum_{x_1,x_2=1}^{n_1,n_2} \left( \|V(x_1,x_2) - D G(x_1,x_2)\|_2^2 + \alpha \|G(x_1,x_2)\|_1 \right)
\end{align} 
with sparsity contraints $\|d^k \|_{2,1} \le \beta$, $k = 1,\ldots,K$, where
\begin{align}
\|d^k\|_{2,1} &\coloneqq \sum_{t=1}^T |\mathbf{d}^k(t)|, \quad |\mathbf{d}^k(t)| 
=  \left( d_1^k(t)^2 + d_2^k(t)^2 \right)^\frac12, \; k=1,\ldots,K.
\end{align}
By reshaping the whole velocity field $\left( \mathbf{v}(t,x_1,x_2) \right)_{t,x_1,x_2}$ into a matrix $V \in \R^{2T, n}$,
this problem can be written in the compact form
\begin{align} \label{eq:model_short} 
\argmin_{D,G} \|V - D G\|_F^2 + \alpha \|G\|_1 \quad \text{subject to} \quad \|d^k \|_{2,1} \le \beta, \; k = 1,\ldots,K,
\end{align} 
where $\|\cdot \|_F$ denotes the Frobenius norm of a matrix and
\begin{align}
\|G\|_1 &\coloneqq \sum_{k=1}^K \|G^k\|_1, \quad 
\|G^k\|_1 \coloneqq \sum_{x_1,x_2=1}^{n_1,n_2} |G^k(x_1,x_2)|.
\end{align}

\begin{remark}\label{rem:compare}
The original sparse dictionary decomposition model in \cite{sparseDictLearning} considers - in our notation -
the minimization problem
\begin{align} \label{eq:model_old} 
\frac12 \|V - D G\|_F^2 + \alpha \|G\|_1 \quad \text{subject to} \quad \|d^k \|_2 \le 1, \; k = 1,\ldots,K
\end{align}
which enforces only the sparsity of the spatial decomposition in $G$. 
In contrast, we are also interested in the (grouped) sparsity of the $d^k$ in time.
We will compare this model with \eqref{eq:model} 
in our numerical examples. %, see Figure 5.
\end{remark}

Problem \eqref{eq:model_short}  is convex with respect to each component $D$ and $G$
and the minimization can be done by alternating with respect to these components.
\\
1.
\textbf{For fixed $D$}, the minimization problem can be separately solved for each $(x_1,x_2)$,
$$
\argmin_{G(x_1,x_2)} \frac12 \|V(x_1,x_2) - D G(x_1,x_2)\|_2^2 + \alpha \|G(x_1,x_2)\|_1, 
\quad x_i \in \{1,\ldots,n_i\}, i=1,2
$$
by several approaches as operator splittings using soft thresholding, 
see  \cite{Bec17,BSS16}
or the LARS method \cite{OPT2000,EHST2004}. 
In this paper, we applied the later one.
\\
2. 
\textbf{For fixed $G$}, we first note that for the trace $\text{tr}$ of a matrix and $A,B \in \R^{M,N}$ it holds
$$
\text{tr} (A^\tT B) = \text{tr} (B A^\tT) = \langle A, B \rangle,
\quad \text{where} \quad \langle A,B \rangle = \sum_{i=1}^{M} \sum_{i=1}^{M}  a_{i,j} b_{i,j}, 
$$
so that we can rewrite the data term in \eqref{eq:model_short} as
\begin{align}
\frac12 \|V - D G\|_2^2 
&= \langle V, V \rangle + \frac12 \langle D G, DG \rangle - \langle V, DG \rangle,\\
&= \text{const} + \frac12 \text{tr}(D^\tT D \underbrace{G G^\tT}_{A \in \R^{K,K}}) - \text{tr}(D^\tT \underbrace{V G^\tT}_{B \in \R^{2T,K}}).
\end{align}
Then we can solve the equivalent problem
\begin{align} \label{eq:model_D_1} 
\argmin_{D} \Big\{\underbrace{\frac12 \text{tr} ( D^\tT D  A)  - \text{tr}(D^\tT B) }_{F(D)} \Big\} \quad \text{subject to} \quad \| d^k \|_{2,1} \le 1, \; k = 1,\ldots,K.
\end{align}
This is done iteratively for each column $d^k$, $k=1,\ldots,K$ of $D$ while keeping the other columns fixed.
Taking the symmetry of $A = (a^1 | \ldots |a^K) = \left(a(j,k) \right)_{j,k=1}^K$ into account, 
we get for the minimizer of $F$ with respect to the $k$-th column of $D$ that
$$
0 = \nabla_{d^k} F(D) = D a^k - b^k \quad \Leftrightarrow \quad d^k = d^k + \frac{1}{a(k,k)} (b^k - D a^k).
$$ 
The algorithm performs one step of the above fixed point iteration to get a new column vector 
$$\widetilde{d^k} = d^k + \frac{1}{a(k,k)} (b^k - D a^k)$$ 
and projects this vector
onto the $(2,1)$-ball given by the $k$th constraint in \eqref{eq:model_D_1}.
This projection $\Pi$ can be done by the so-called grouped shrinkage which reads for 
$\widetilde{d^k} = \left( \widetilde{\mathbf{d}^k}(t) \right)_{t=1}^T$ as
$$
\Pi\left(\widetilde{\mathbf{d}^k}(t) \right) 
= 
\left\{
\begin{array}{ll}
\widetilde{\mathbf{d}^k}(t)& \text{if } |\widetilde{\mathbf{d}^k}(t)| \le \beta,\\
\beta \widetilde{\mathbf{d}^k}(t)/|\widetilde{\mathbf{d}^k}(t)| & \text{otherwise}.
\end{array}
\right.  \quad t=1,\ldots,T
$$
Finally, the above steps are \emph{not} performed for all spatial points at the same time, but
the sum in \eqref{eq:model}  is updated point by point and the dictionary $D$ from the previous step is used for a ,,warm start''
in the next step. 
This kind of algorithms is known as block-coordinate descent algorithm with warm restarts
see \cite{Ber1999}.  Convergence of the algorithm was shown  under 
certain assumptions in \cite{sparseDictLearning} which can be also applied for our modified setting.
All steps are summarized in Algorithm \ref{algo:2-grid-iter}. 

\begin{algorithm}[tbh]
\caption{Double Sparse OF Decomposition } \label{algo:2-grid-iter}
\begin{algorithmic}
\STATE \textbf{Input:} $V$, $\alpha$, $\beta$
\STATE $A_0 \gets 0$, $B_0 \gets 0$, $D_0$ initial dictionary
\FOR {$r=1$ to $n$}
\STATE draw $(x_1,x_2)$ at random and set $V_r \gets V(x_1,x_2)$ (see \ref{decomp})
\STATE
\STATE Compute $G_r$ using LARS:
\STATE \vspace*{5pt}
$\qquad G_r \gets \argmin_{G(x_1,x_2)} \frac12 \|V_r - D_{r-1} G(x_1,x_2)\|_2^2 + \alpha \|G(x_1,x_2)\|_1$ \vspace*{5pt}
\STATE $A_r \gets A_{r-1} + G_r G_r^\tT$
\STATE $B_r \gets B_{r-1} + V_r G_r^\tT$
\STATE
\STATE Update $D$ columnwise by:
\FOR {$k=1$ to $K$} 
\STATE $A \gets A_r$
\STATE $B \gets B_r$
\STATE $D \gets D_{r-1}$
\STATE Update the $k$-the column of $D$ by 
\STATE $\widetilde{d^k} \gets \frac{1}{a(k,k)}(b^k - D a^k) + d^k$
\STATE $\mathbf{d}^k(t) = \left\{
\begin{array}{ll}
\widetilde{\mathbf{d}^k}(t)& \text{if } |\widetilde{\mathbf{d}^k}(t)| \le \beta,\\
\beta \widetilde{\mathbf{d}^k}(t)/|\widetilde{\mathbf{d}^k}(t)| & \text{otherwise}.
\end{array}
\right. \quad t=1,\ldots,T
$
\STATE $d^k \gets \left(\mathbf{d}^k(t)\right)_{t=1}^T$
\ENDFOR
\ENDFOR
\end{algorithmic}
\end{algorithm}

%------------------------------------------------------------
\section{Motion Magnification} \label{sec:MM}
%------------------------------------------------------------

The detection of microexpressions and facial micro movements plays an increasingly important role for various applications from computer vision. Microexpressions and micro movements are often unobservable by the {untrained} and naked eye. This difficult task can be necessary in the context of the annotation of video datasets (e.g. compare \cite{yan2014casme}). The visualization of microexpressions and other micro movements in the face via motion magnification can potentially mitigate the difficulty of manual microexpression detection to enable manual annotation and microexpression analysis without professional or specific training. Furthermore, unsupervised selection and magnification of different motion components can help with exploratory investigation of facial micro movements for the development of new methods for remote assessment.

Having decomposed the OF field into meaningful spatial and temporal components,
we show in this section how these components can be used to enhance the OF in the regions of interest
and how this can be visualized in a new video sequence using a sophisticated warping method. Image warping is used in the context of OF estimation, compression or image registration (usually backwards warping) or for applications from graphics such as texture mapping or novel view synthesis, among the rich literature, see, e.g. \cite{brox2004high,wohlberg}.

\subsection{Forward Warping}
%As we aim to warp the image frames in order to amplify micro-motions, 
%we need to take a closer look at the used technique. 
For a given displacement $v$ between two consecutive image frames $f_1$ and $f_2$ (so that we can skip the time variable), there are two prevalent methods for warping, forward warping and backward warping:
\begin{itemize} 
\item[i)] Backward warping $f_1$ to yield $f_{\text{bw}}$:
\begin{align*}
    f_{\text{bw}} (x_1, x_2) \coloneqq f_1(x_1 - v_1(x_1,x_2), x_2 - v_2(x_1,x_2)) .
\end{align*}
Usually backward warping is preferred as it  directly computes the warped frame.
However, we will see that for warping facial motions, backward warping is far inferior to forward warping.
\item[ii)] Forward warping $f_1$ to yield $f_{\text{fw}}$: 
\begin{align*}
    f_{\text{fw}} \left(x_1 + v_1(x_1,x_2), x_2 + v_2(x_1,x_2) \right) \coloneqq f_1(x_1, x_2).
\end{align*}
\end{itemize}

\begin{figure}
    \centering
    \includegraphics[scale=0.45]{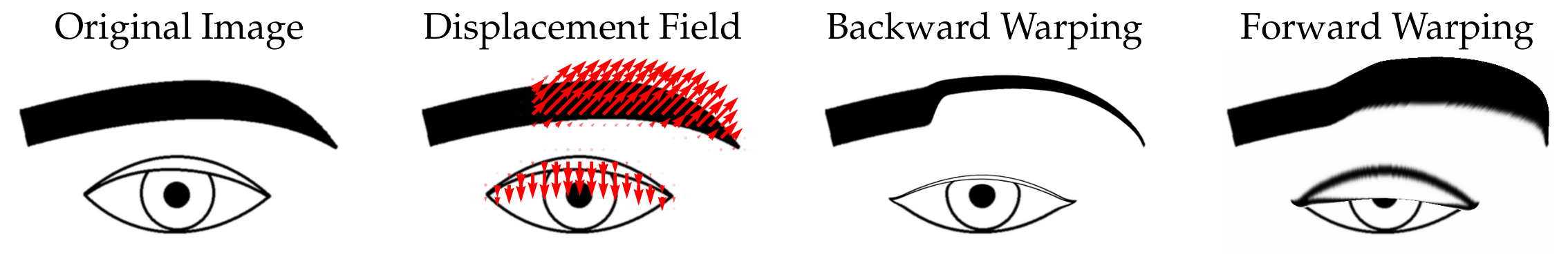}
    \caption{An illustration showing the results of forward versus backward warping with an example displacement field.}
    \label{fig:forward_backward_warping}
\end{figure}

The difference between  both warping methods  is illustrated in Figure \ref{fig:forward_backward_warping}. This figure shows the results of forward warping versus backward warping on a displacement field which corresponds to an {magnified} blinking motion as well as a {magnification of the} raising of an eyebrow. In this case, the motion field is localized on the eyebrow and the eyelid. Therefore, when doing backward warping, the colors stay exactly the same everywhere the displacement field is zero. In this illustration, this yields a contraction of the eyebrow and a contraction of the eye itself. This problem does not occur when doing forward warping.

\begin{figure}
    \centering
    \includegraphics[scale=0.4]{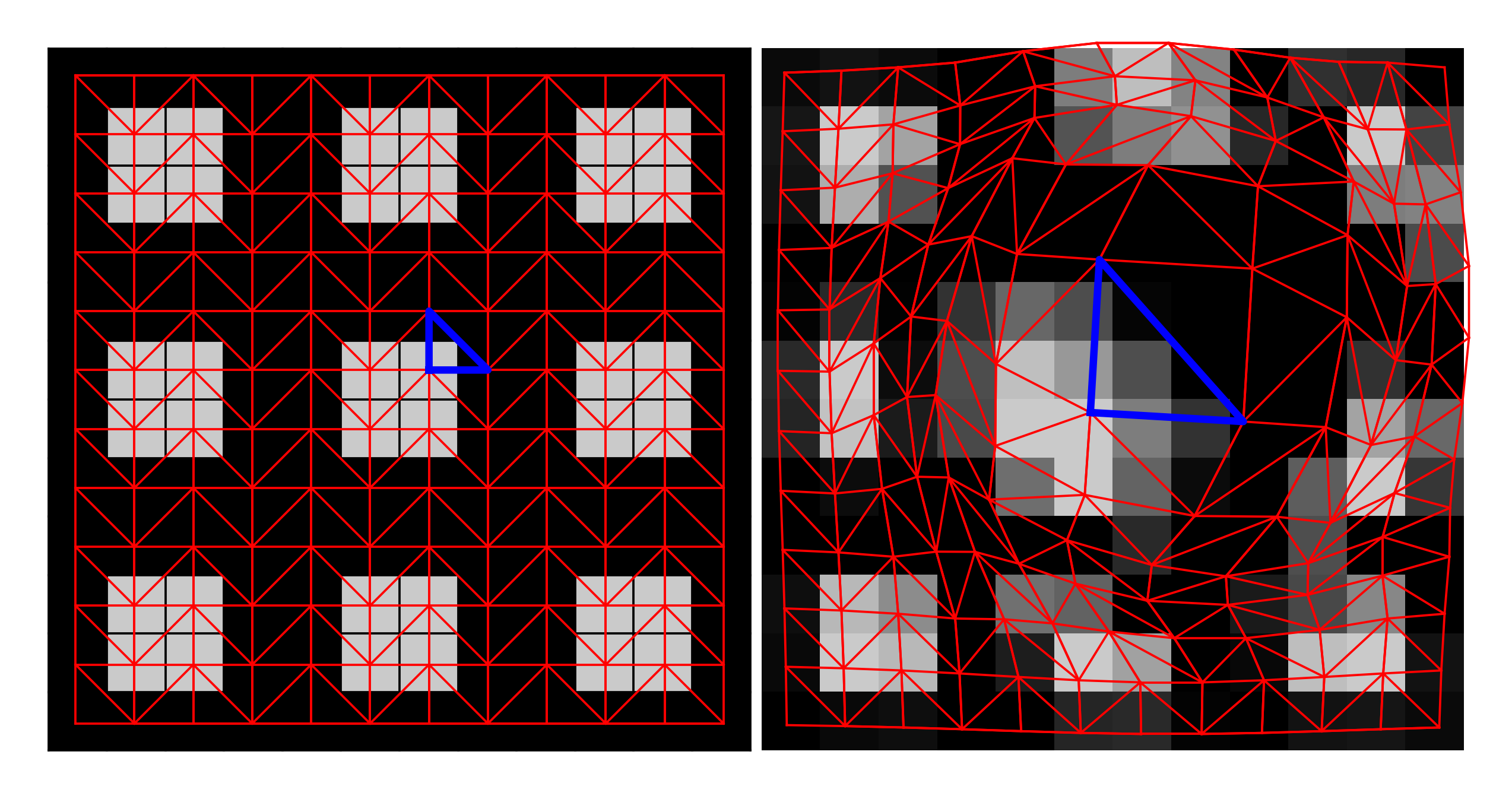}
    \caption{Illustration of the forward warping process. The image is first triangulated as shown on the left. The triangles are displaced and the warped image is then rasterized from these distorted triangles, see next figure.}
    \label{fig:forward_warping_fig}
\end{figure}

However, in computer graphics, forward warping is significantly harder to do. This is because for the warped frame, we only have color information at the points $(x_1 + v_1(x_1,x_2), x_2 + v_2(x_1,x_2))$, 
which generally do not correspond to grid coordinates.
To overcome this problem, we use the following scheme: 
first, we consider a triangulation of the frame $f_1$, which we aim to warp. The structure of this triangulation is shown in Figure~\ref{fig:forward_warping_fig} left. Each triangle connects three adjacent pixels and
the vertices are equipped with the color values of the corresponding pixel. 
Then the triangles are displaced according to the vector field $(v_1, v_2)$ yielding a displaced triangulation
depicted in Figure~\ref{fig:forward_warping_fig} right. 
Finally, each of the displaced triangles is rasterized. 
More precisely, we take all pixels whose midpoints are contained in the displaced triangle and assign them a value interpolating the triangle vertex color values using barycentric coordinates.

Finally, it can be possible that multiple displaced triangles overlap. To resolve this ambiguity, the proposed algorithm allows for a depth map as input which indicates which triangles should be drawn above others. While for this application, an elaborate depth map based on facial features could be devised, we opted to set the depth map such that stronger motions always ``overlap'' weaker motions.

To efficiently compute the triangulation, displacement and subsequent rasterization of the input image, we employ a shader pipeline written using the OpenGL programming language. In short, this means that we make use of GPU native routines which are extremely efficient at handling such tasks and enable us to even employ this forward warping technique in real-time applications.

\subsection{Motion Magnification in Selected Regions}
With all this at hand, we are now able to {magnify} selected movements in the input video. To this end, we still need to decide which components from the decomposition in Section~\ref{sec:PCA} can be considered to be micro-movements. In our case, this is done by simply evaluating for each $k\in \{1,\ldots,K\}$, if the values of $d^k$ lie in a specified range. However, as the $d^k$ are all normalized due to our optimization procedure, we need to adjust them to better resemble the velocity magnitudes from the OF. In short, we do the following selection:
\begin{enumerate}
    \item Compute the maximum of each domain $G^k$ for all $k\in \{1,\ldots,K\}$,
    \begin{align*}
        m_k \coloneqq \max_{x_1, x_2} G^k(x_1,x_2).
    \end{align*}
    \item Compute the maximal normalized motion magnitude for each component
    \begin{align*}
        c^k \coloneqq m^k\max_{t\in \{1,\ldots,T\}} \norm{d^k(t)}_2.
    \end{align*}
    \item For two pre-defined thresholds $0 < \lambda_1 < \lambda_2$, consider the $k$-th component $k\in \{1,\ldots,K\}$ to belong to a micro-movement if 
		$\lambda_1 \leq c^k \leq \lambda_2.$
\end{enumerate}

Let $\mathcal{I} \subset \{1,\ldots,K\}$ be the set of components deemed to correspond to micro-movements by the above selection. Then, we {magnify} their motion as follows. For each, $t\in \{1,\ldots,T-1\}$, we assume that $\bfv(t, \cdot)$ warps the frame $f_1$ into $f_2$. Therefore, {magnification of} the micro-movements by the {magnification} factor $\mu > 0$, we aim to warp $f_1$ by the flow field
\begin{align*}
    \tilde{\bfv}(t, \cdot) \coloneqq \bfv(t, \cdot) + \mu \sum_{k\in \mathcal{I}} d^k(t) G^k(\cdot).
\end{align*}

\begin{figure}[h!]
    \centering
    \includegraphics[width=0.75\linewidth]{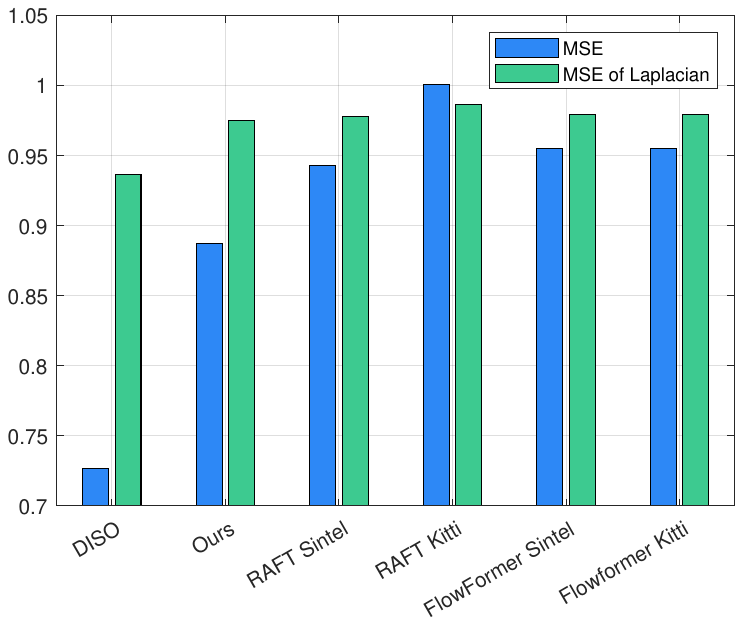}
    \caption{MSE for a sequence on a microexpression sequence after motion compensation. The values are normalized with respect to the MSE of the raw recording. While DISO performs better than the refined RAFT, our RAFT refinement on the ME dataset improved the performance over other state-of-the-art models trained on Sintel and KITTI-2015 in our examples. }
    \label{fig:mse}
\end{figure}

\section{Experiments}\label{sec:numerics}
\begin{figure}
\centering
\includegraphics[width=\linewidth]{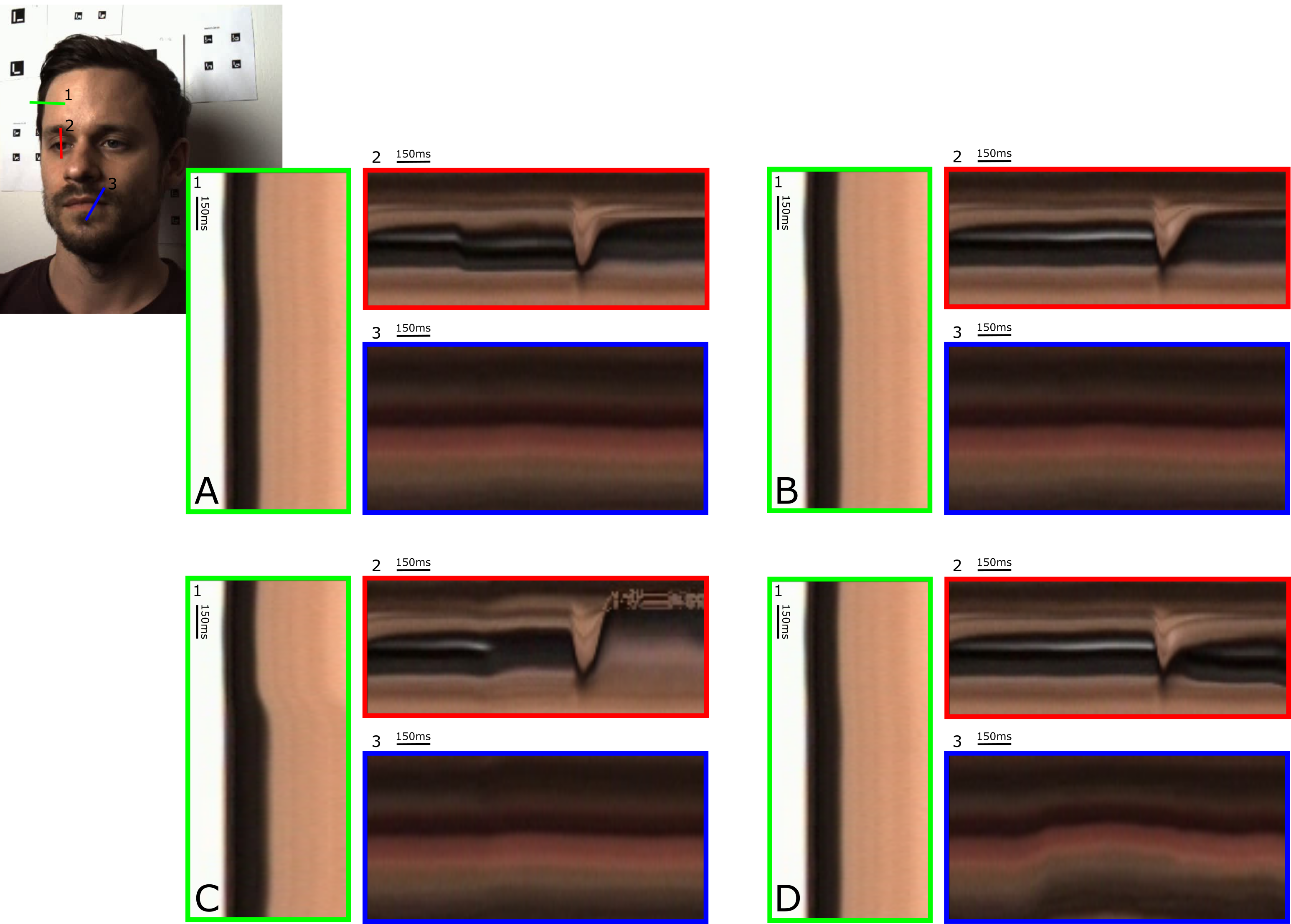}
\caption{Experiments on the medical dataset with different parameters. The choice of parameters can be used to visualize different motion components when analysing the video.
(A,2): {magnification} of a subtle saccade before the blinking event with
$(\eta,\lambda,\mu) = (3,0.2,10)$;
(B,2): {magnification} of a larger saccade after blinking
with $(\eta,\lambda,\mu) = (3,0.4,4)$; 
(C,1): {magnification} the subtle head motion with 
$(\eta,\lambda,\mu) = (4,0.5,8)$;
(D,3): {magnification} of a micromovement of the mouth with 
$(\eta,\lambda,\mu) = (8,0.3,10)$. All results are with $K=9$ and $\lambda_1=0.1$.}
\label{fig:vi-screen}
\end{figure}

%------------------------------------------------------------
We illustrate our proposed method on an example video from CASME II \cite{yan2014casme} and a video from the medical dataset for the remote detection of respiratory infections (see Figure \protect\ref{fig:vi-screen}). 
All our experiments {were performed on consumer-grade computers}. RAFT training has been performed on a multi-GPU machine with Intel Xeon Gold processors and two Nvidia Tesla V100 GPUs.

\paragraph{Datasets and data acquisition.} CASME II and SMIC are datasets of spontaneous microexpressions which include 35 (CASME II) and 16 (SMIC) participants respectively. Videos are recorded at 200fps (CASME II) and 100fps (SMIC). Each video is labelled with either observable or expected microexpressions, the latter based on self-reports of the participants. 
% vi-screen:
Additionally, we use a video  from a dataset for the remote assessment of respiratory diseases (VI-SCREEN). For this dataset, participants were recruited at multiple recording sites, including Saarland univerisity clinic (UKS) Children's Hospital and at the UKS central emergency admission, Germany. Recruitment happened from February 2022 and is ongoing (July 2023). Up until April 2023, this study includes 106 participants (median age 26 years, \textit{interquartile range} (IQR) 24 years). Informed consent was obtained, and the study was approved by the Ethics committee of the Ärztekammer des Saarlandes (02/22). Viral and bacterial, respiratory pathogens were tested using multiplex polymerase chain reaction (Multiplex PCR) assays. 
The camera array consisted of stereo RGB, near infrared (NIR) and thermal cameras as well as microphones (compare \cite{flotho2021multimodal}). A duration between 5-10 minutes was recorded where participants answered questions in front of the camera. Core temperature was recorded with infrared ear thermometers. The videos in this dataset can contain microexpressions as well as other types of micro movements which are potentially related to symptoms of respiratory diseases. As benchmark sequence, we selected a video which contains a large movement (eye blinking) as well as a small movement around the mouth. 

\subsection{Motion estimation}
We compare our method against RAFT-Sintel and RAFT-KITTI as well as against FlowFormer\footnote{FlowFormer is among the top ten best performing methods on the Sintel benchmark \cite{Butler:ECCV:2012} as of July 2023.} after the same training schedules \mbox{\cite{huang2022flowformer}}.
RAFT-Sintel performed better than RAFT-KITTI
which is why we choose RAFT-Sintel as baseline model for refinement. 
Due to the lack of publicly available datasets 
of facial microexpressions with ground truth OF annotation, 
we compared the methods {by estimating a motion compensation error on a CASME II sequence from participant number 14\footnote{The benchmark video is \texttt{EP04\_04f} from participant number 14.}.
We used the \textit{mean squared error} (MSE) of the gray values 
as well as on the Laplacian}, see Figure~\ref{fig:mse}. 
While the performance of DISO 
was better on the test sequence than all other methods, 
our RAFT refinement with the ME dataset could generally improve performance on facial microexpression videos.  

\subsection{Motion magnification}
For the motion magnification experiments, we use a video of participant number 14 from CASME II {as well as a recording 
from the VI-SCREEN dataset. Participant number 14 has been} excluded from the OF training dataset.
The videos contain microexpressions and micromovements.
We perform the analysis outlined in Section~\ref{sec:PCA} 
using a sparsity regularization parameters $\alpha = 0.1$, $\beta = 4$ and  $K=9$ components. 
A motion component $k \in \{1,\ldots, 9\}$ 
is selected to be a microexpression if $\lambda_1 = 0.1 \leq c^k \leq \lambda_2 = 0.3$ holds true. 
We {magnify} these selected microexpressions using the magnification factor $\mu=4$.

\begin{figure}
    \centering
    \includegraphics[scale=0.16]{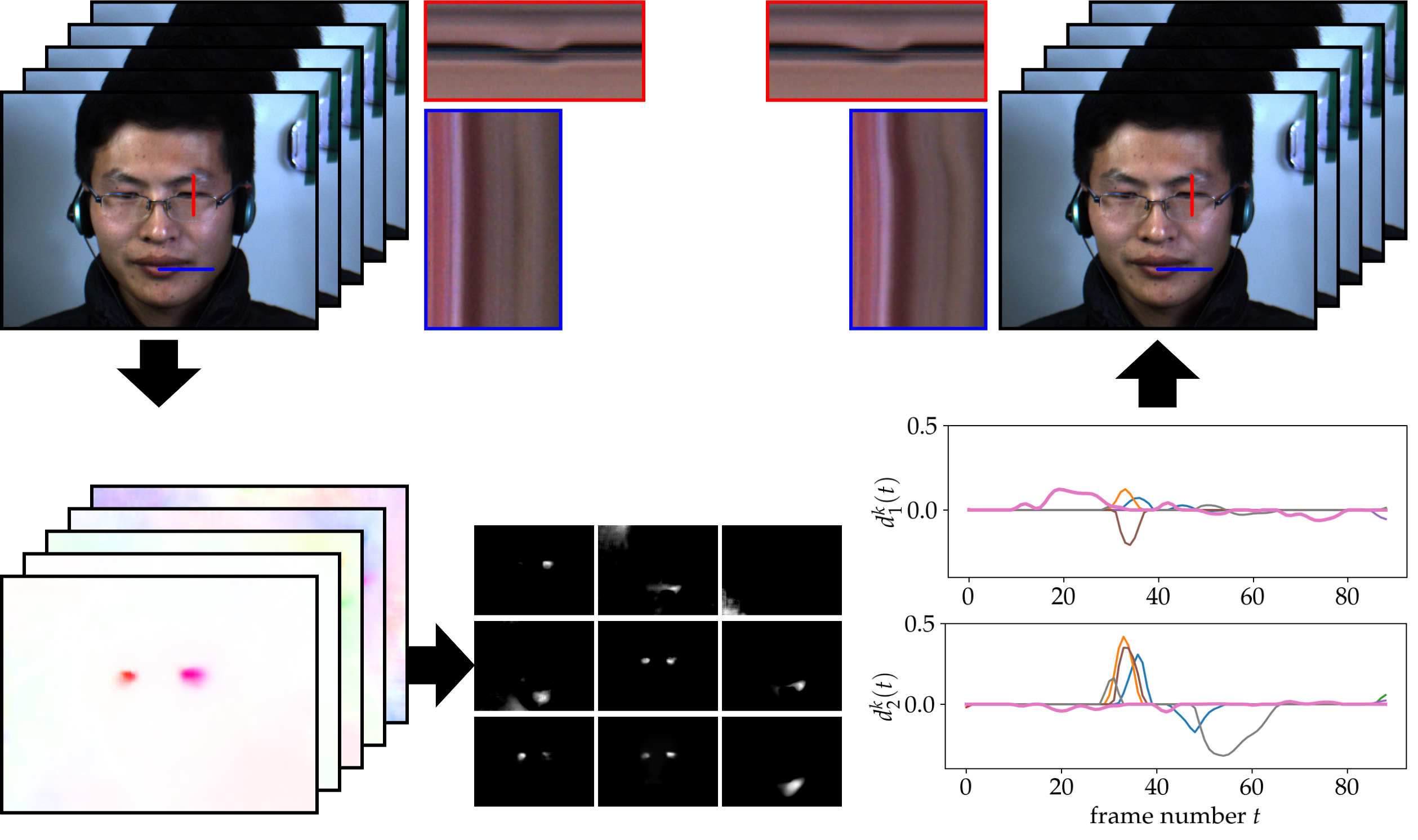}
    \caption{Illustration of our motion magnification procedure 
		with the $\norm{\cdot}_{2,1}$-constraint. The algorithm starts with a sequence of $80$ 
		video frames of size $640 {\times} 480$ illustrated in the top left. 
		The OF field computed for each time step is illustrated in the lower left. 
		Then we use our method to decompose this OF field into $G^1,\ldots,G^9$ 
		shown in the bottom middle 
		as well as the $(d^1_1,d^1_2),\ldots,(d^9_1, d^9_2)$ in the bottom right. 
		In this example, two components, shown by thicker lines in the plot, 
		were selected by thresholding to be micromovements. The frames are warped accordingly 
		to yield the video sequence in the top right. 
		The pixels from the red and blue lines are plotted over time 
		in the corresponding red and blue boxes. This shows that the motion around the mouth 
		is  {magnified} while leaving the blinking motion unchanged.}
    \label{fig:diagram_12_norm}
\end{figure}

\begin{figure}
    \centering
    \includegraphics[scale=0.32]{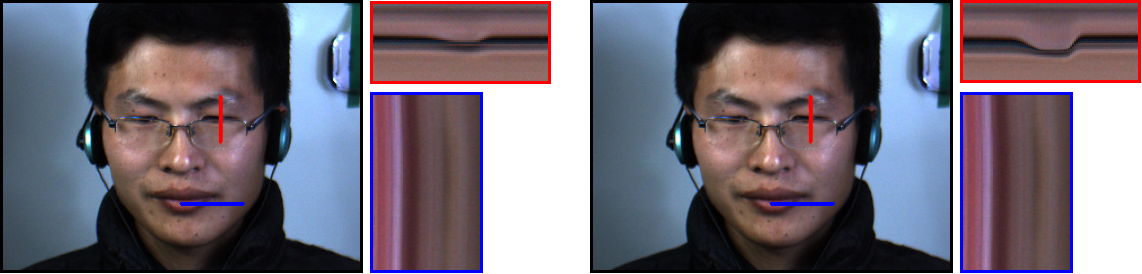}
    \caption{Illustration of the same video {without (left) and with (right) motion magnification. The chosen} thresholds {($\lambda_1 = 0.3, \lambda_2 = \infty$)} only magnify the largest motions {}: This results in { magnification of the} blinking motion {while} leaving other facial movements unchanged.
    }
    \label{fig:eye_magnification}
\end{figure}

\begin{figure}
    \centering
    \includegraphics[scale=0.16]{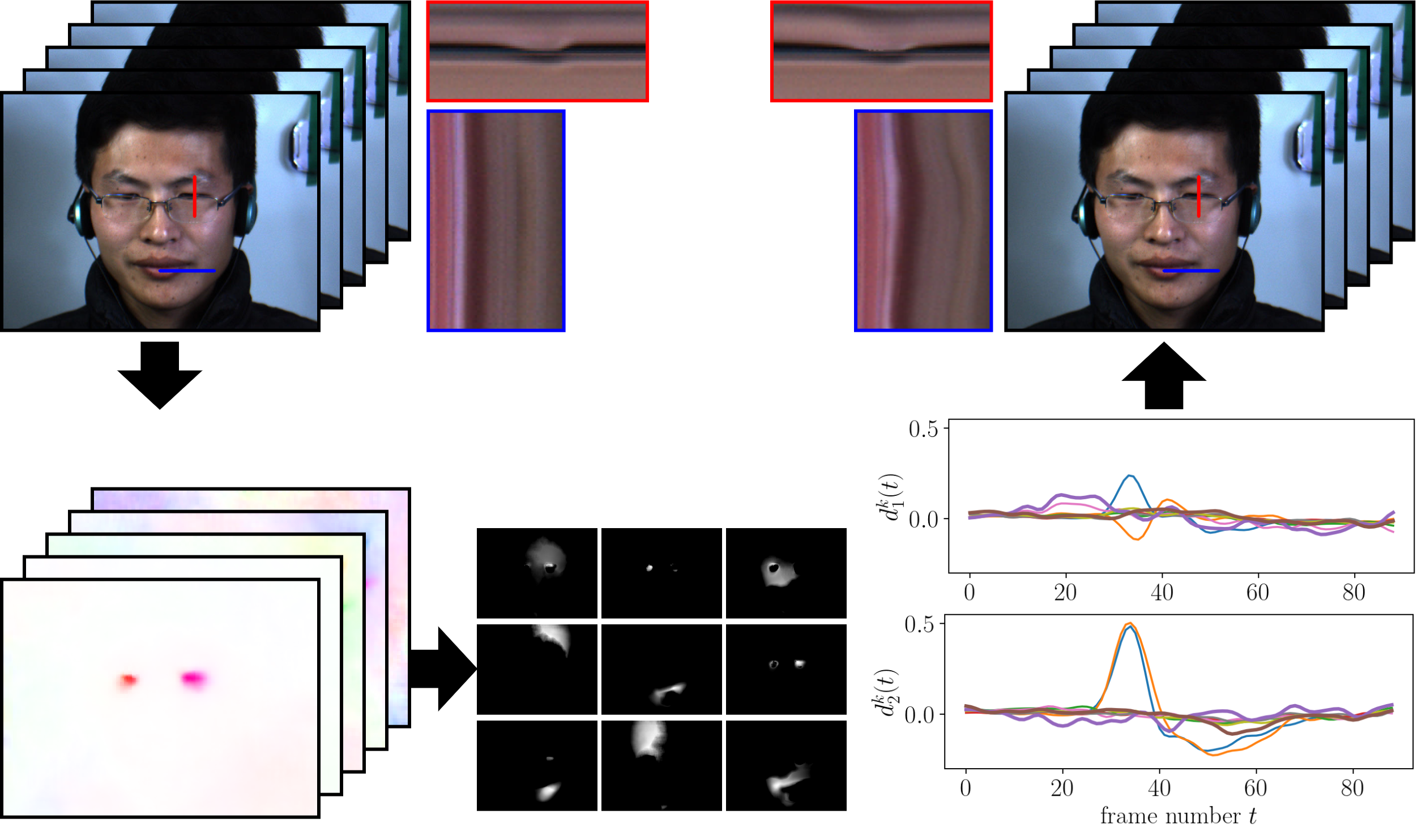}
    \caption{Illustration of our motion magnification procedure as in 
		Figure \ref{fig:diagram_12_norm}, but with 
		$\norm{\cdot}_{2}$-constraint.}
    \label{fig:diagram_old}
\end{figure}

{Figure~\ref{fig:diagram_12_norm} shows an example application of our method with the sparse decomposition model \eqref{eq:model_short} with $\norm{\cdot}_{2,1}$ constraint.}
Our experiments show that our method is able to successfully decompose the OF field into reasonable components. 
Using this decomposition, we are able to perform facial motion magnification selectively in an unsupervised manner. 
For the case shown in Figure~\ref{fig:diagram_12_norm} and Figure~\ref{fig:diagram_old}, the threshold corresponds only to the components connected to the micromovements around the mouth. The blinking motion is unchanged. Changing the thresholds also lets us select different motions. In Figure~\ref{fig:eye_magnification}, we illustrate the results of our algorithm when choosing instead the thresholds $\lambda_1 = 0.3$ and $\lambda_2=\infty$. In this case, only the blinking motion is  {magnified}.

\begin{figure}
    \centering
    \includegraphics[scale=1]{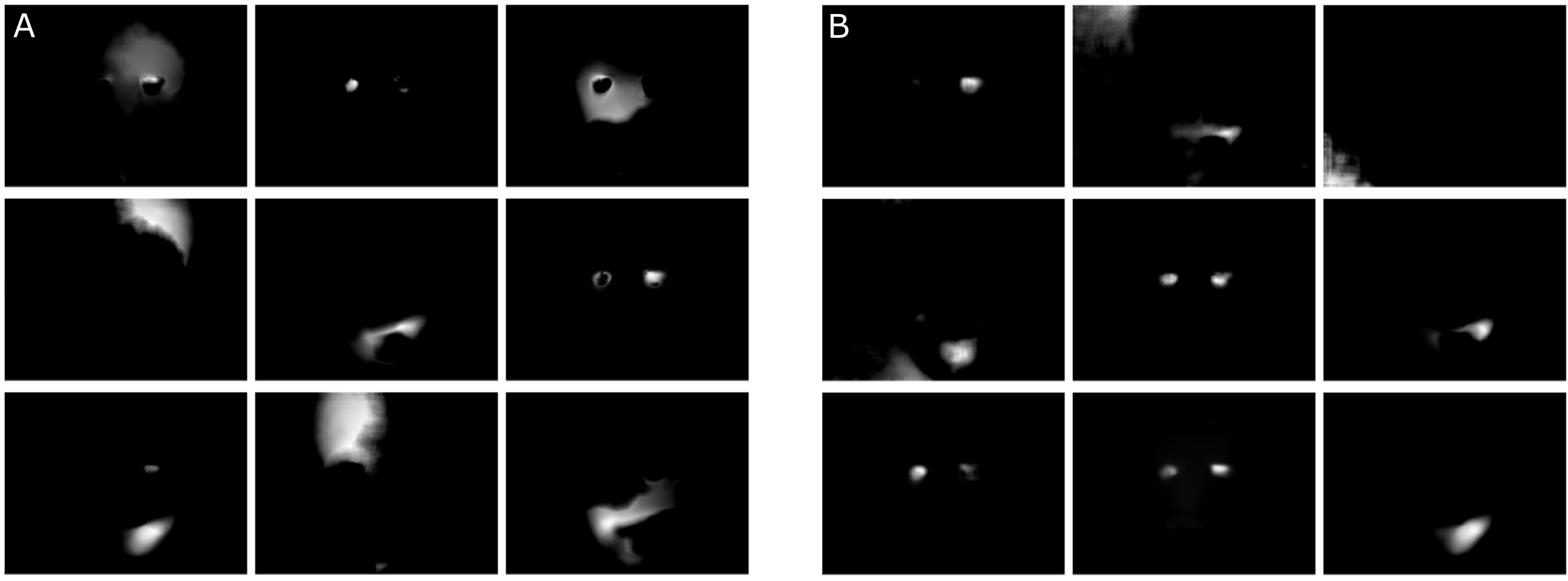}
    \caption{Illustration of the motion components $G^1,\ldots,G^9$ with 
		$\norm{\cdot}_{2}$-constraint (A) and with the $\norm{\cdot}_{2,1}$-constraint (B). The components in (B) show spatially smaller components with clearer edges. }
    \label{fig:components}
\end{figure}

\begin{figure}
    \centering
    \includegraphics[width=\linewidth]{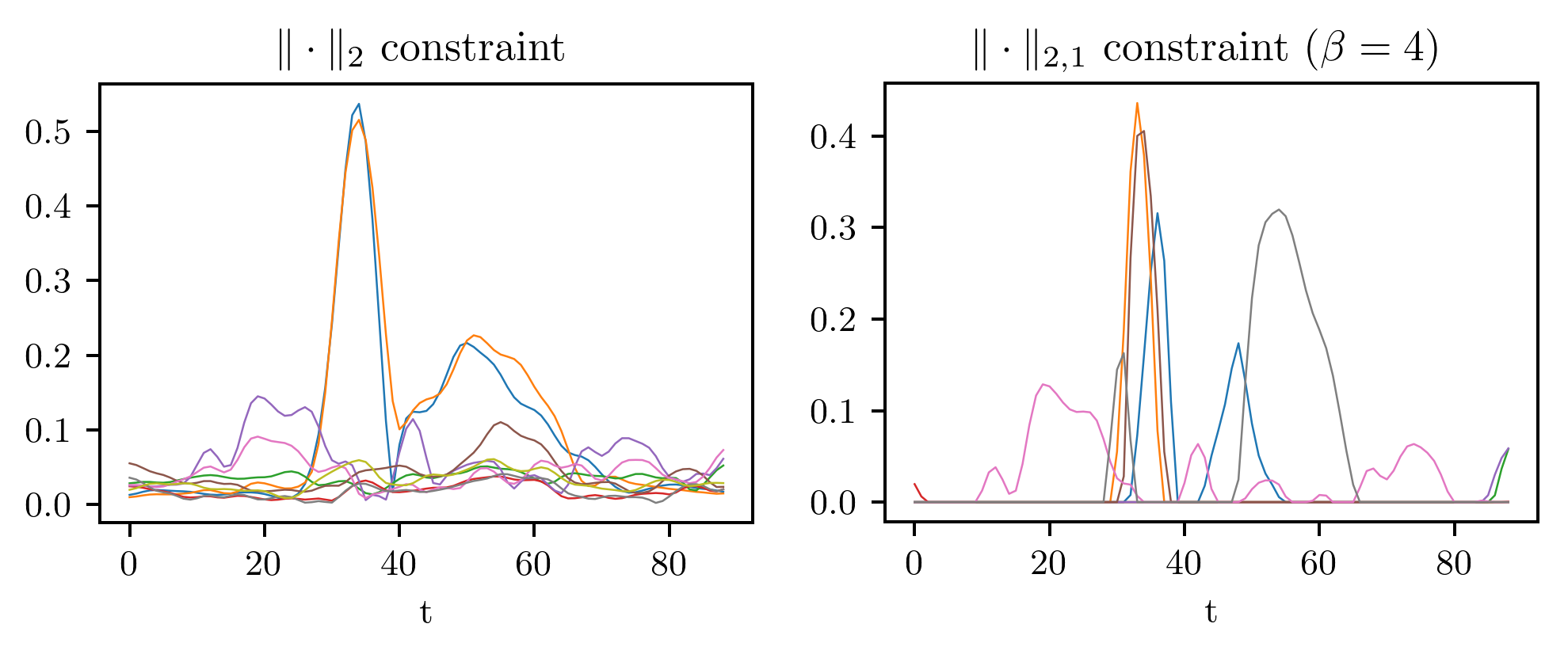}
    \caption{On the left-hand side, $\Vert d^k(t)\Vert_2$ is shown over time for all nine components, where we optimized  problem~\eqref{eq:model_old}.
    On the right-hand side, we see the curves of $\| d^k(t) \|_{2,1}$ after optimizing problem~\eqref{eq:model_short}. It can be observed that the later one promotes the separation of the flow field in temporally distinct motions, while in the first one   a few components capture a wide range of motion events.}
    \label{fig:code_sparsity_comparison}
\end{figure}

For comparison, Figure~\ref{fig:diagram_old} depicts the outcome of a similar experiment using just the $\norm{\cdot}_2$-constraint,
see Remark \ref{rem:compare}. For this comparison, we use the same parameters $\alpha=0.1$, $K=9$ and $\mu = 4$.
Introduction of the sparsity-promoting constraint leads to a clearer separation of motion components in the image as well as between temporally disconnected motions. We can observe this in the visualization of the components in Figure~\ref{fig:components}, where the $\norm{\cdot}_{2,1}$-constraint produces smaller components with clearer edges. Figure~\ref{fig:code_sparsity_comparison} shows how each component from the $\norm{\cdot}_{2,1}$-constraint has only limited, temporal extend, while with the $\norm{\cdot}_{2}$-constraint a few components capture a wide range of motion events over the complete video.

\par

%, \alpha=0.15, \mu=8$)

%------------------------------------------------------------

%------------------------------------------------------------
\section{Conclusions} \label{sec:conclusions}

We  provided a non-supervised method for magnifying micro-motions in facial videos
based on the Lagrangian approach with OF forward warping {magnification}.
The actual regions for OF {magnification} were found by minimizing a double sparse decomposition 
model for OF and a thresholding procedure to detect the relevant regions.
Our thresholding method for differentiating microexpressions from other movements is open to further improvement. 
For example, the temporal extent of each motion could be taken into consideration. However, in case of the CASME II dataset, where eye movements are the only notable facial motions beside microexpressions, a simple thresholding was already sufficient. Furthermore, significant head-movements might disrupt the presented decomposition into time-independent regions of motion in the face. In these cases, a facial alignment pre-processing step might be necessary. This could be realized similar to related work\mbox{\cite{flotho2022lagrangian, flotho2018lagrangian, elgharib2015video}} or by attenuating large and global motion with our method which might become more important with out-of-lab data with difficult motion. Also, our method has multiple parameters that need to be manually chosen. As a next step, statistics from real world microexpression datasets could be included to learn suitable parameters for different motion types. The presented methods have the potential to facilitate the annotation of datasets with such face recordings for microexpression detection. Subtle expression changes, where the detection would otherwise require expert training, could be magnified for untrained individuals, who could then annotate such data. Our preliminary results with a real world, medical recording demonstrated how different parameters would allow for the visualization of different motion components in videos.

Up to our knowledge, there exist no facial microexpression or micromovement datasets with ground truth OF. OF methods have been analysed with respect to their predictive power of microexpressions \mbox{\cite{allaert2022comparative}} which does not allow conclusions on the overall quality of the OF predictions. Regarding the refinement of RAFT, future evaluations have to show the performance gain compared to OF methods tailored to faces with different training or transfer learning strategies such as \cite{alkaddour2021self} for which the models are currently not publicly available. 
Given the small movements in the SMIC and CASME II recordings as well as constant backgrounds, further data augmentations that introduce large displacements and different background textures could further improve the method.

\subsection{Acknowledgements}
This work has partially been funded by Federal Ministry of Education and Research (BMBF, grant numbers 13N15753 and 13N15754). 

\bibliographystyle{abbrv}
\bibliography{Bibliography}

\begin{thebibliography}{10}

\bibitem{alinovi2018respiratory}
D.~Alinovi, G.~Ferrari, F.~Pisani, and R.~Raheli.
\newblock Respiratory rate monitoring by video processing using local motion
  magnification.
\newblock In {\em Proc Eur Signal Process Conf EUSIPCO}, pages 1780--1784.
  IEEE, 2018.

\bibitem{alkaddour2021self}
M.~Alkaddour, U.~Tariq, and A.~Dhall.
\newblock Self-supervised approach for facial movement based optical flow.
\newblock {\em arXiv}, 2021.

\bibitem{allaert2022comparative}
B.~Allaert, I.~R. Ward, M.~Bilasco, C.~Djeraba, and M.~Bennamoun.
\newblock A comparative study on optical flow for facial expression analysis.
\newblock {\em Neurocomputing}, 2022.

\bibitem{bai2012selectively}
J.~Bai, A.~Agarwala, M.~Agrawala, and R.~Ramamoorthi.
\newblock Selectively de-animating video.
\newblock {\em ACM Trans Graph}, 31(4):66--1, 2012.

\bibitem{baker2011database}
S.~Baker, D.~Scharstein, J.~P. Lewis, S.~Roth, M.~J. Black, and R.~Szeliski.
\newblock A database and evaluation methodology for optical flow.
\newblock {\em Int J Comput Vis}, 92(1):1--31, 2011.

\bibitem{Bec17}
A.~Beck.
\newblock {\em First-Order Methods in Optimization}.
\newblock Society for Industrial and Applied Mathematics (SIAM), Mathematical
  Optimization Society, Philadelphia, 2017.

\bibitem{BePeSch15}
F.~Becker, S.~Petra, and C.~Schn{\"o}rr.
\newblock Optical flow.
\newblock In {\em Handbook of Mathematical Methods in Imaging}, pages
  1945--2004. Springer, New York, 2015.

\bibitem{BS2021}
R.~Beinert and G.~Steidl.
\newblock Robust {{PCA}} via regularized reaper with a matrix-free proximal
  algorithm.
\newblock {\em Journal of Mathematical Imaging and Vision}, pages 626--649,
  2021.

\bibitem{Ber1999}
D.~P. Bertsekas.
\newblock {\em Nonlinear Programming}.
\newblock Athena Scientific Belmont, 1999.

\bibitem{opencv_library}
G.~Bradski.
\newblock {The OpenCV Library}.
\newblock {\em Dr. Dobb's Journal of Software Tools}, 2000.

\bibitem{brox2004high}
T.~Brox, A.~Bruhn, N.~Papenberg, and J.~Weickert.
\newblock High accuracy optical flow estimation based on a theory for warping.
\newblock In {\em Comput Vis ECCV}, pages 25--36. Springer, 2004.

\bibitem{BSS16}
M.~Burger, A.~Sawatzky, and G.~Steidl.
\newblock First order algorithms in variational image processing.
\newblock In R.~Glowinski, S.~J. Osher, and W.~Yin, editors, {\em Splitting
  Methods in Communication, Imaging, Science, and Engineering}, pages 345--407.
  Springer, Cham, 2016.

\bibitem{Butler:ECCV:2012}
D.~J. Butler, J.~Wulff, G.~B. Stanley, and M.~J. Black.
\newblock A naturalistic open source movie for optical flow evaluation.
\newblock In {A. Fitzgibbon et al. (Eds.)}, editor, {\em European Conf. on
  Computer Vision (ECCV)}, Part IV, LNCS 7577, pages 611--625. Springer-Verlag,
  Oct. 2012.

\bibitem{CTCB2013}
R.~Cabral, F.~D. la~Torre, J.~Costeira, and A.~Bernardino.
\newblock Unifying nuclear norm and bilinear factorization approaches for
  low-rank matrix decomposition.
\newblock In {\em Proceedings of the 2013 IEEE International Conference on
  Computer Vision}, 2013.

\bibitem{CLMW2011}
E.~Cand\'es, X.~Li, Y.~Ma, and J.~Wright.
\newblock Robust principal component analysis?
\newblock {\em Journal of the ACM}, 58(3):1--37, 2011.

\bibitem{chen2013large}
Z.~Chen, H.~Jin, Z.~Lin, S.~Cohen, and Y.~Wu.
\newblock Large displacement optical flow from nearest neighbor fields.
\newblock In {\em Proc IEEE Comput Soc Conf Comput Vis Pattern Recognit}, pages
  2443--2450, 2013.

\bibitem{DFIB15}
A.~Dosovitskiy, P.~Fischer, E.~Ilg, P.~H{\"a}usser, C.~Haz{\i}rba{\c{s}},
  V.~Golkov, P.~v.d. Smagt, D.~Cremers, and T.~Brox.
\newblock Flownet: Learning optical flow with convolutional networks.
\newblock In {\em IEEE International Conference on Computer Vision (ICCV)},
  2015.

\bibitem{EHST2004}
B.~Efron, T.~Hastie, I.~Johnstone, and R.~Tibshirani.
\newblock Least angle regression.
\newblock {\em Ann Stat}, 32(2):407--499, 2004.

\bibitem{eitner2021modal}
M.~Eitner, M.~Musta, L.~Vanstone, J.~Sirohi, and N.~Clemens.
\newblock Modal parameter estimation of a compliant panel using phase-based
  motion magnification and stereoscopic digital image correlation.
\newblock {\em Experimental Techniques}, 45(3):287--296, 2021.

\bibitem{elgharib2015video}
M.~Elgharib, M.~Hefeeda, F.~Durand, and W.~T. Freeman.
\newblock Video magnification in presence of large motions.
\newblock In {\em Proc IEEE Comput Soc Conf Comput Vis Pattern Recognit}, pages
  4119--4127, 2015.

\bibitem{fei2021exposing}
J.~Fei, Z.~Xia, P.~Yu, and F.~Xiao.
\newblock Exposing ai-generated videos with motion magnification.
\newblock {\em Multimed Tools Appl}, 80(20):30789--30802, 2021.

\bibitem{flotho2021multimodal}
P.~Flotho, M.~J. Bhamborae, T.~Gr{\"u}n, C.~Trenado, D.~Thinnes, D.~Limbach,
  and D.~J. Strauss.
\newblock Multimodal data acquisition at sars-cov-2 drive through screening
  centers: Setup description and experiences in saarland, germany.
\newblock {\em J. Biophotonics}, 14(8):e202000512, 2021.

\bibitem{flotho2018lagrangian}
P.~Flotho, M.~J. Bhamborae, L.~Haab, and D.~J. Strauss.
\newblock Lagrangian motion magnification revisited: Continuous, magnitude
  driven motion scaling for psychophysiological experiments.
\newblock In {\em Annu Int Conf IEEE Eng Med Biol Soc}, pages 3586--3589. IEEE,
  2018.

\bibitem{flotho2022lagrangian}
P.~Flotho, C.~Hei{\ss}, G.~Steidl, and D.~J. Strauss.
\newblock Lagrangian motion magnification with landmark-prior and sparse pca
  for facial microexpressions and micromovements.
\newblock In {\em Annu Int Conf IEEE Eng Med Biol Soc}, pages 2215--2218. IEEE,
  2022.

\bibitem{flotho2022a}
P.~Flotho, S.~Nomura, B.~Kuhn, and D.~J. Strauss.
\newblock Software for non-parametric image registration of 2-photon imaging
  data.
\newblock {\em J Biophotonics}, page e202100330, 2022.
\newblock e202100330 jbio.202100330.R2.

\bibitem{hartmann2021measurement}
C.~Hartmann, H.~A. Weiss, P.~Lechner, W.~Volk, S.~Neumayer, J.~H. Fitschen, and
  G.~Steidl.
\newblock Measurement of strain, strain rate and crack evolution in shear
  cutting.
\newblock {\em J Mater Process Technol}, 288:116872, 2021.

\bibitem{HK2014}
M.~Holler and K.~Kunisch.
\newblock On infimal convolution of tv-type functionals and applications to
  video and image reconstruction.
\newblock {\em SIAM J. Imag. Sci.}, 7(4):2258--2300, 2014.

\bibitem{HS81}
B.~K. Horn and B.~G. Schunck.
\newblock {Determining optical flow}.
\newblock {\em Art Intelligence}, 17(1-3):185--203, 1981.

\bibitem{huang2017discriminative}
X.~Huang, S.-J. Wang, X.~Liu, G.~Zhao, X.~Feng, and M.~Pietik{\"a}inen.
\newblock Discriminative spatiotemporal local binary pattern with revisited
  integral projection for spontaneous facial micro-expression recognition.
\newblock {\em IEEE Transactions on Affective Computing}, 10(1):32--47, 2017.

\bibitem{huang2022flowformer}
Z.~Huang, X.~Shi, C.~Zhang, Q.~Wang, K.~C. Cheung, H.~Qin, J.~Dai, and H.~Li.
\newblock Flowformer: A transformer architecture for optical flow.
\newblock {\em arXiv preprint arXiv:2203.16194}, 2022.

\bibitem{kondermann2016hci}
D.~Kondermann, R.~Nair, K.~Honauer, K.~Krispin, J.~Andrulis, A.~Brock,
  B.~Gussefeld, M.~Rahimimoghaddam, S.~Hofmann, C.~Brenner, and B.~Jähne.
\newblock The hci benchmark suite: Stereo and flow ground truth with
  uncertainties for urban autonomous driving.
\newblock In {\em Proceedings of the IEEE Conference on Computer Vision and
  Pattern Recognition Workshops}, pages 19--28, 2016.

\bibitem{kroeger2016fast}
T.~Kroeger, R.~Timofte, D.~Dai, and L.~Van~Gool.
\newblock Fast optical flow using dense inverse search.
\newblock In {\em Comput Vis ECCV}, pages 471--488. Springer, 2016.

\bibitem{li2013spontaneous}
X.~Li, T.~Pfister, X.~Huang, G.~Zhao, and M.~Pietik{\"a}inen.
\newblock A spontaneous micro-expression database: Inducement, collection and
  baseline.
\newblock In {\em IEEE Int Conf Autom Face Gesture Recognit Workshops}, pages
  1--6. IEEE, 2013.

\bibitem{FirstMotionMag}
C.~Liu, A.~Torralba, W.~T. Freeman, F.~Durand, and E.~H. Adelson.
\newblock Motion magnification.
\newblock {\em SIGGRAPH}, 24(3), 2005.

\bibitem{sparseDictLearning}
J.~Mairal, F.~Bach, J.~Ponce, and G.~Sapiro.
\newblock Online dictionary learning for sparse coding.
\newblock In {\em Annu Int Conf on Mach Learn}, ICML '09, page 689–696, New
  York, NY, USA, 2009. Association for Computing Machinery.

\bibitem{mayer2016large}
N.~Mayer, E.~Ilg, P.~Hausser, P.~Fischer, D.~Cremers, A.~Dosovitskiy, and
  T.~Brox.
\newblock A large dataset to train convolutional networks for disparity,
  optical flow, and scene flow estimation.
\newblock In {\em Proc IEEE Comput Soc Conf Comput Vis Pattern Recognit}, pages
  4040--4048, 2016.

\bibitem{menze2015object}
M.~Menze and A.~Geiger.
\newblock Object scene flow for autonomous vehicles.
\newblock In {\em Proceedings of the IEEE conference on computer vision and
  pattern recognition}, pages 3061--3070, 2015.

\bibitem{mroueh2015deep}
Y.~Mroueh, E.~Marcheret, and V.~Goel.
\newblock Deep multimodal learning for audio-visual speech recognition.
\newblock In {\em Proc IEEE Int Conf Acoust Speech Signal Process}, pages
  2130--2134. IEEE, 2015.

\bibitem{le2018micro}
L.~N., A.~C., J.~A., P.~R. C.-W., and S.~J.
\newblock Micro-expression motion magnification: Global {L}agrangian vs. local
  {E}ulerian approaches.
\newblock In {\em IEEE Int Conf Autom Face Gesture Recognit Workshops}, pages
  650--656. IEEE, 2018.

\bibitem{OPT2000}
M.~R. Osborne, B.~Presnell, and B.~A. Turlach.
\newblock A new approach to variable selection in least squares problems.
\newblock {\em IMA J Numer Anal}, 20(3):389--403, 2000.

\bibitem{sarrafi2018vibration}
A.~Sarrafi, Z.~Mao, C.~Niezrecki, and P.~Poozesh.
\newblock Vibration-based damage detection in wind turbine blades using
  phase-based motion estimation and motion magnification.
\newblock {\em J Sound Vib}, 421:300--318, 2018.

\bibitem{shan2013video}
L.~Shan and M.~Yu.
\newblock Video-based heart rate measurement using head motion tracking and
  ica.
\newblock In {\em 2013 6th International Congress on Image and Signal
  Processing (CISP)}, volume~1, pages 160--164. IEEE, 2013.

\bibitem{strauss2020vestigial}
D.~J. Strauss, F.~I. Corona-Strauss, A.~Schroeer, P.~Flotho, R.~Hannemann, and
  S.~A. Hackley.
\newblock Vestigial auriculomotor activity indicates the direction of auditory
  attention in humans.
\newblock {\em Elife}, 9:e54536, 2020.

\bibitem{sun2022makes}
D.~Sun, C.~Herrmann, F.~Reda, M.~Rubinstein, D.~J. Fleet, and W.~T. Freeman.
\newblock Disentangling architecture and training for optical flow.
\newblock In {\em European Conf. on Computer Vision (ECCV)}, pages 165--182.
  Springer, 2022.

\bibitem{sun2010secrets}
D.~Sun, S.~Roth, and M.~J. Black.
\newblock Secrets of optical flow estimation and their principles.
\newblock In {\em 2010 IEEE computer society conference on computer vision and
  pattern recognition}, pages 2432--2439. IEEE, 2010.

\bibitem{teed2020raft}
Z.~Teed and J.~Deng.
\newblock Raft: Recurrent all-pairs field transforms for optical flow.
\newblock In {\em Comput Vis ECCV}, pages 402--419. Springer, 2020.

\bibitem{wadhwa2013phase}
N.~Wadhwa, M.~Rubinstein, F.~Durand, and W.~T. Freeman.
\newblock Phase-based video motion processing.
\newblock {\em ACM Trans Graph}, 32(4):80, 2013.

\bibitem{wohlberg}
G.~Wolberg.
\newblock Digital image warping.
\newblock 1990.

\bibitem{wu2012eulerian}
H.-Y. Wu, M.~Rubinstein, E.~Shih, J.~Guttag, F.~Durand, and W.~Freeman.
\newblock Eulerian video magnification for revealing subtle changes in the
  world.
\newblock {\em ACM Trans Graph}, 31(4):1--8, 2012.

\bibitem{xu2017microexpression}
F.~Xu, J.~Zhang, and J.~Z. Wang.
\newblock Microexpression identification and categorization using a facial
  dynamics map.
\newblock {\em IEEE Trans Affect Comput}, 8(2):254--267, 2017.

\bibitem{yan2014casme}
W.-J. Yan, X.~Li, S.-J. Wang, G.~Zhao, Y.-J. Liu, Y.-H. Chen, and X.~Fu.
\newblock Casme ii: An improved spontaneous micro-expression database and the
  baseline evaluation.
\newblock {\em PLoS One}, 9(1):e86041, 2014.

\bibitem{yan2013fast}
W.-J. Yan, Q.~Wu, J.~Liang, Y.-H. Chen, and X.~Fu.
\newblock How fast are the leaked facial expressions: The duration of
  micro-expressions.
\newblock {\em J Nonverbal Behav}, 37(4):217--230, 2013.

\bibitem{zhang2021review}
L.~Zhang and O.~Arandjelovi{\'c}.
\newblock Review of automatic microexpression recognition in the past decade.
\newblock {\em Mach Learn Knowl Extr}, 3(2):414--434, 2021.

\bibitem{zhang2017video}
Y.~Zhang, S.~L. Pintea, and J.~C. Van~Gemert.
\newblock Video acceleration magnification.
\newblock In {\em Proc IEEE Comput Soc Conf Comput Vis Pattern Recognit}, pages
  529--537, 2017.

\end{thebibliography}

\end{document}